\documentstyle[aclap,fleqn,psfig]{article}
\input{ipamacs}

\font\ipatenrm=wsuipa10

\title{Linguistic Structure as Composition and Perturbation}
\author{Carl de Marcken \\
MIT AI Laboratory, NE43-769 \\
545 Technology Square \\
Cambridge, MA, 02139, USA \\{\bf cgdemarc@ai.mit.edu}}

\begin{document}
\setlength{\baselineskip}{.15in}
\maketitle

\begin{abstract}
  This paper discusses the problem of learning language from unprocessed
  text and speech signals, concentrating on the problem of learning a
  lexicon.  In particular, it argues for a representation of language in
  which linguistic parameters like words are built by perturbing a
  composition of existing parameters.  The power of this representation is
  demonstrated by several examples in text segmentation and compression,
  acquisition of a lexicon from raw speech, and the acquisition of mappings
  between text and artificial representations of meaning. 
\end{abstract}

\newcommand{\pageline}{\underline{\hspace{\columnwidth}}}
\newcommand{\word}[1]{{\em #1}}
\newcommand{\sword}[2]{{\em #1} {$\{${\sc #2}$\}$}}
\newcommand{\arpB}{b}
\newcommand{\arpD}{d}
\newcommand{\arpG}{g}
\newcommand{\arpP}{p}
\newcommand{\arpT}{t}
\newcommand{\arpK}{k}
\newcommand{\arpBCL}{b$^{\mbox{\tiny cl}}$}
\newcommand{\arpDCL}{d$^{\mbox{\tiny cl}}$}
\newcommand{\arpGCL}{g$^{\mbox{\tiny cl}}$}
\newcommand{\arpPCL}{p$^{\mbox{\tiny cl}}$}
\newcommand{\arpTCL}{t$^{\mbox{\tiny cl}}$}
\newcommand{\arpKCL}{k$^{\mbox{\tiny cl}}$}
\newcommand{\arpDX}{\flapr}
\newcommand{\arpQ}{\glotstop}
\newcommand{\arpJH}{\v{j}}
\newcommand{\arpCH}{\v{c}}
\newcommand{\arpS}{s}
\newcommand{\arpSH}{\v{s}}
\newcommand{\arpZ}{z}
\newcommand{\arpZH}{\v{z}}
\newcommand{\arpF}{f}
\newcommand{\arpTH}{\nitheta}
\newcommand{\arpV}{v}
\newcommand{\arpDH}{\eth}
\newcommand{\arpM}{m}
\newcommand{\arpN}{n}
\newcommand{\arpNG}{\eng}
\newcommand{\arpEM}{{\undercirc m}}
\newcommand{\arpEN}{{\undercirc n}}
\newcommand{\arpENG}{{\undercirc \eng}}
\newcommand{\arpNX}{\eng}
\newcommand{\arpL}{l}
\newcommand{\arpR}{r}
\newcommand{\arpW}{w}
\newcommand{\arpY}{y}
\newcommand{\arpHH}{h}
\newcommand{\arpHV}{\hookh}
\newcommand{\arpEL}{{\undercirc l}}

\newcommand{\arpIY}{i}
\newcommand{\arpIH}{\sci}
\newcommand{\arpEH}{\niepsilon}
\newcommand{\arpEY}{e$^{\mbox{\tiny y}}$}
\newcommand{\arpAE}{\ae}
\newcommand{\arpAA}{a}
\newcommand{\arpAW}{\scripta$^{\mbox{\tiny w}}$}
\newcommand{\arpAY}{\scripta$^{\mbox{\tiny y}}$}
\newcommand{\arpAH}{\invv}
\newcommand{\arpAO}{\openo}
\newcommand{\arpOY}{\openo$^{\mbox{\tiny y}}$}
\newcommand{\arpOW}{o}
\newcommand{\arpUH}{\niupsilon}
\newcommand{\arpUW}{u}
\newcommand{\arpUX}{\"{u}}
\newcommand{\arpER}{\hookrevepsilon}
\newcommand{\arpAX}{\schwa}
\newcommand{\arpIX}{\bari}
\newcommand{\arpAXR}{\er}
\newcommand{\arpAXH}{{\undercirc \er}}

\newlength{\pausewidth}
\settowidth{\pausewidth}{i}
\newlength{\longpausewidth}
\settowidth{\longpausewidth}{w}

\newcommand{\arpPAU}{\hspace{\longpausewidth}}
\newcommand{\arpEPI}{-}

\newcommand{\unib}{b}
\newcommand{\unid}{d}
\newcommand{\unig}{g}
\newcommand{\unip}{p}
\newcommand{\unit}{t}
\newcommand{\unik}{k}
\newcommand{\uniJ}{\v{j}}
\newcommand{\uniC}{\v{c}}
\newcommand{\unis}{s}
\newcommand{\uniS}{\v{s}}
\newcommand{\uniz}{z}
\newcommand{\uniZ}{\v{z}}
\newcommand{\unif}{f}
\newcommand{\uniT}{\nitheta}
\newcommand{\univ}{v}
\newcommand{\uniD}{\eth}
\newcommand{\unim}{m}
\newcommand{\unin}{n}
\newcommand{\uniG}{\eng}
\newcommand{\unil}{l}
\newcommand{\unir}{r}
\newcommand{\uniw}{w}
\newcommand{\uniy}{y}
\newcommand{\unih}{h}
\newcommand{\uniH}{\hookh}
\newcommand{\unii}{i}
\newcommand{\uniI}{\sci}
\newcommand{\uniE}{\niepsilon}
\newcommand{\unie}{e}
\newcommand{\uniA}{\ae}
\newcommand{\unia}{a}
\newcommand{\uniAH}{\invv} 
\newcommand{\uniO}{\openo}
\newcommand{\unio}{o}
\newcommand{\uniU}{\niupsilon}
\newcommand{\uniu}{u}
\newcommand{\uniAX}{\schwa} 
\newcommand{\uniIX}{\bari} 
\newcommand{\uniEPI}{-} 

\section{Motivation}

Language is a robust and necessarily redundant communication mechanism. 
Its redundancies commonly manifest themselves as predictable patterns in
speech and text signals, and it is largely these patterns that enable text
and speech compression.  Naturally, many patterns in text and speech
reflect interesting properties of language.  For example, \word{the} is
both an unusually frequent sequence of letters and an English word.  This
suggests using compression as a means of acquiring underlying properties of
language from surface signals.  The general methodology of
language-learning-by-compression is not new.  Some notable early proponents
included Chomsky~\shortcite{Chomsky55}, Solomonoff~\shortcite{Solomonoff60}
and Harris~\shortcite{Harris68}, and compression has been used as the basis
for a wide variety of computer programs that attack unsupervised learning
in language; see
\cite{Olivier68,Wolff82,Ellison92,Stolcke94,Chen95,Cartwright94} among
others. 

\subsection{Patterns and Language}

Unfortunately, while surface patterns often reflect interesting linguistic
mechanisms and parameters, they do not always do so.  Three classes of
examples serve to illustrate this. 

\subsubsection{Extralinguistic Patterns}\label{extralinguistic}

The sequence \word{it was a dark and stormy night} is a pattern in the
sense it occurs in text far more often than the frequencies of its
letters would suggest, but that does not make it a lexical or grammatical
primitive: it is the product of a complex mixture of linguistic and
extra-linguistic processes.  Such patterns can be indistinguishable from
desired ones.  For example, in the Brown corpus~\cite{Francis82}
\word{scratching her nose} occurs 5 times, a corpus-specific idiosyncrasy. 
This phrase has the same structure as the idiom \word{kicking the bucket}. 
It is difficult to imagine any induction algorithm learning \word{kicking
  the bucket} from this corpus without also (mistakenly) learning
\word{scratching her nose}. 

\subsubsection{The Definition of Interesting}\label{definteresting}

This discussion presumes there is a set of desired patterns to extract from
input signals.  What is this set?  For example, is \word{kicking the
  bucket} a proper lexical unit?  The answer depends on factors external to
the unsupervised learning framework.  For the purposes of machine
translation or information retrieval this sequence is an important idiom,
but with respect to speech recognition it is unremarkable.  Similar
questions could be asked of subword units like syllables.  Plainly, the
answers depends on the learning context, and not on the signal itself. 

\subsubsection{The Definition of Pattern}\label{defpattern}

Any statistical definition of pattern depends on an underlying model.  For
instance, the sequence \word{the dog} occurs much more frequently than one
would expect given an independence assumption about letters.  But for a
model with knowledge of syntax and word probabilities, there is nothing
remarkable about the phrase.  Since all existing models have flaws,
patterns will always appear that are artifacts of imperfections in the
learning algorithm. 

These examples seem to imply that unsupervised induction will never
converge to ideal grammars and lexicons.  While there is truth to this, the
rest of this paper describes a representation of language that bypasses
many of the apparent difficulties. 

\section{A Compositional Representation}

The examples in sections~\ref{extralinguistic} and \ref{definteresting}
seem to imply that any unsupervised language learning program that returns
only one interpretation of the input is bound to make many mistakes.  And
section~\ref{defpattern} implies that decisions about linguistic units must
be made relative to their representations.  Both of these issues are
addressed if linguistic units (for now, words in the lexicon) are built by
composing other units.  For example, \word{kicking the bucket} might be
represented by the composition of \word{kicking}, \word{the} and
\word{bucket}.\footnote{The simplest composition operator is concatenation;
  sections~\ref{meaning} and \ref{grammar} discuss more interesting ones.}
Of course, words that are merely the composition of their parts are
uninteresting and need not be included in the lexicon.  The motivation for
including a word in the lexicon must be that it behaves differently than
its parts imply.  If this is the case, a word is a perturbation of a
composition.

In the case of \word{kicking the bucket} the perturbation is one of both
meaning and frequency.  For \word{scratching her nose} the perturbation may
just be of frequency.\footnote{Naturally, an unsupervised learning
  algorithm with no access to meaning will not treat these two examples
  differently.} This is a very natural representation from the viewpoint of
language.  It correctly predicts that both phrases inherit their sound and
syntax from their component words.  At the same time it leaves open the
possibility that idiosyncratic information will be attached to the whole,
as with the meaning of \word{kicking the bucket}.  This structure is very
much like the class hierarchy of a modern programming language.  It is not
the same thing as a context-free grammar, since each word does not act in
the same way as the default composition of its components. 

\begin{figure}[t]
\pageline
\begin{center}
\mbox{\ }
\psfig{figure=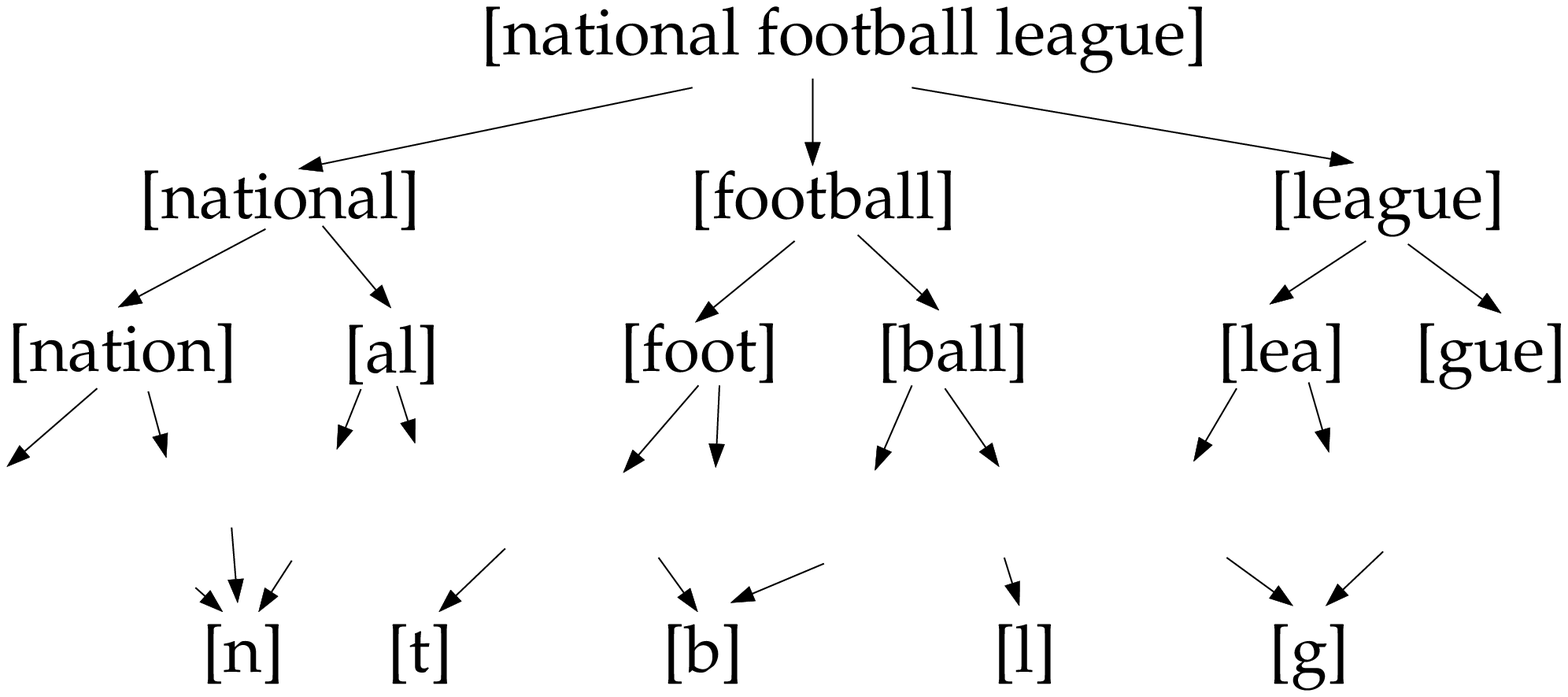,height=1.5in,width=3.0in}
\end{center}

\caption{\label{fig:comp}A compositional representation.}
\pageline
\end{figure}

Figure~\ref{fig:comp} illustrates a recursive decomposition (under
concatenation) of the phrase \word{national football league}.  The phrase
is broken into three words, each of which are also decomposed in the
lexicon.  This process bottoms out in the terminal characters.  This is a
real decomposition achieved by a program described in section~\ref{concat}. 
Not shown are the perturbations (in this case merely probability specifications)
that distinguish each word from its parts.  This general framework extends
to other perturbations.  For example, the word \word{wanna} is naturally
thought of as a composition of \word{want} and \word{to} with a sound
change.  And in speech the three different words \word{to}, \word{two} and
\word{too} may well inherit the sound of a common ancestor while
introducing new syntactic and semantic properties. 

\subsection{Coding}

Of course, for this representation to be more than an intuition both the
composition and perturbation operators must be exactly specified.  In
particular, a code must be designed that enables a word (or the input) to
be expressed in terms of its parts.  As a simple example, suppose that the
composition operator is concatenation, that terminals are characters, and
that the only perturbation operator is the ability to express the
probability of a word independently of the probability of its parts.  Then
to code either the input or a (nonterminal) word in the lexicon, the number
of component words in the representation is written, followed by a
code for each component word.  Naturally, each word in the lexicon must
also be linked to its code, and under a near-optimal coding scheme like a
Huffman code, the code length will be related to the probability of the
word.  Thus, linking a word to a code serves also to specify the word's
probability, its only perturbation.  Furthermore, if words are written down
in order of decreasing probability, a Huffman code for a large lexicon can
be specified using a negligible number of bits (providing the number of
codes of each length is sufficient).  This and the near-negligible cost of
writing down the number of components in word representations will not be
discussed further.  Figure~\ref{fig:code} presents a portion of an encoding
of a hypothetical lexicon under this scheme.

\begin{figure}[t]
\pageline
\begin{center}
\begin{tabular}{lcl}
Code & Length & Components \\ \hline
$000\ (= c_{\mbox{\tiny of}})$ & 2 & $c_{\mbox{\tiny o}}, c_{\mbox{\tiny f}}$ \\
$001\ (= c_{\mbox{\tiny the}})$ & 3 & $c_{\mbox{\tiny t}}, c_{\mbox{\tiny h}}, c_{\mbox{\tiny e}}$ \\
$010\ (= c_{\mbox{\tiny in}})$ & 2 & $c_{\mbox{\tiny i}}, c_{\mbox{\tiny n}}$ \\
$0110\ (= c_{\mbox{\tiny some}})$ & 4 & $c_{\mbox{\tiny s}}, c_{\mbox{\tiny o}}, c_{\mbox{\tiny m}}, c_{\mbox{\tiny e}}$ \\
$0111\ (= c_{\mbox{\tiny someofthe}})$ & 3 & $c_{\mbox{\tiny some}}, c_{\mbox{\tiny of}}, c_{\mbox{\tiny the}}$ \\
$10000\ \ldots$ & \ldots & \ldots \\
\end{tabular}
\end{center}
\caption{\label{fig:code}A coding of the first few words of a
  hypothetical lexicon.  The first two columns can be coded succinctly,
  leaving the cost of pointers to component words as the dominant cost of
  both the lexicon and the representation of the input.} \pageline
\end{figure}

\subsection{MDL}

Given a coding scheme and a particular lexicon (and a parsing algorithm) it
is in theory possible to calculate the minimum length encoding of a
given input.  Part of the encoding will be devoted to the lexicon, the rest
to representing the input in terms of the lexicon.  The lexicon that
minimizes the combined description length of the lexicon and the input
maximally compresses the input.  In the sense of Rissanen's minimum
description-length (MDL) principle \cite{Rissanen78,Rissanen89} this
lexicon is the theory that best explains the data, and one can hope that
the patterns in the lexicon reflect the underlying mechanisms and
parameters of the language that generated the input. 

\subsection{Properties of the Representation}

Representing words in the lexicon as perturbations of compositions has a
number of desirable properties.  

\begin{itemize}

\item The choice of composition and perturbation operators captures
  a particular detailed theory of language.  They can be used, for instance,
  to reference sophisticated phonological and morphological mechanisms. 

\item The length of the description of a word is a measure of its
  linguistic plausibility, and can serve as a buffer against learning
  unnatural coincidences. 

\item Coincidences like \word{scratching her nose} do not exclude
  desired structure, since they are further broken down
  into components that they inherit properties from. 

\item Structure is shared: the words \word{blackbird} and \word{blackberry}
  can share the common substructure associated with \word{black}, such as
  its sound and meaning.  As a consequence, data is pooled for estimation,
  and representations are compact.

\item Common irregular forms are compiled out.  For example, if \word{went}
  is represented in terms of \word{go} (presumably to save the cost of
  unnecessarily reproducing syntactic and semantic properties) the complex
  sound change need only be represented once, not every time \word{went} is
  used.

\item Since parameters (words) have compact representations, they are cheap
  from a description length standpoint, and many can be included in the
  lexicon.  This allows learning algorithms to fit detailed statistical
  properties of the data.

\end{itemize}

This coding scheme is very similar to that found in popular
dictionary-based compression schemes like LZ78 \cite{Ziv78}.  It is capable
of compressing a sequence of identical characters of length $n$ to size
${\cal O}(\log n)$.  However, in contrast to compression schemes like LZ78
that use deterministic rules to add parameters to the dictionary (and do
not arrive at linguistically plausible parameters), it is possible to
perform more sophisticated searches in this representation.

\section{A Search Algorithm}

Since the class of possible lexicons is infinite, the minimization of
description length is necessarily inexact and heuristic.  Given a fixed
lexicon, the expectation-maximization algorithm \cite{Dempster77} can be
used to arrive at a (locally) optimal set of probabilities and codelengths
for the words in the lexicon.  For composition by concatenation, the
algorithm reduces to the special case of the Baum-Welch procedure
\cite{Baum70} discussed in \cite{Deligne95}.  In general, however, the
parsing and re-estimation involved in EM can be considerably more
complicated.  To update the structure of the lexicon, words can be added or
deleted from it if this is predicted to reduce the description length of
the input.  This algorithm is summarized in
figure~\ref{fig:alg}.\footnote{For the composition operators and test sets
  we have looked at, using single (Viterbi) parses produces almost exactly
  the same results (in terms of both compression and lexical structure) as
  summing probabilities over multiple parses.}

\begin{figure}[th]
\pageline
\begin{tabbing}
Start with lexicon of terminals.\\
Ite\=rate\\
\> Ite\=rate (EM)\\
\> \> Parse input and words using current lexicon.\\
\> \> Use word counts to update probabilities.\\
\> Add words to the lexicon.\\
\> Ite\=rate (EM)\\
\> \> Parse input and words using current lexicon.\\
\> \> Use word counts to update probabilities.\\
\> Delete words from the lexicon.
\end{tabbing}
\caption{\label{fig:alg}An iterative search algorithm.  Two iterations of
  the inner loops are usually sufficient for convergence, and for the tests
  described in this paper after 10 iterations of the outer loop there is
  little change in the lexicon in terms of either compression performance
  or structure.  This algorithm is quite practical for the sizes of
  problems presented in this paper.} \pageline
\end{figure}

\subsection{Adding and Deleting Words}

For words to be added to the lexicon, two things are needed.  The first is
a means of hypothesizing candidate new words.  The second is a means of
evaluating candidates.  One reasonable means of generating candidates is to
look at pairs (or bigger tuples) of words that are composed in the parses
of words and the input.  So long as the composition operator is
associative, a new word can be created from such a pair and substituted in
place of it wherever it appears.  For example, if \word{water} and
\word{melon} are frequently composed, then a good candidate for a new word
is \word{water} $\circ$ \word{melon} = \word{watermelon}, where $\circ$ is
the composition operator.  In order to evaluate whether the addition of
such a new word is likely to reduce the description length of the input, it
is necessary to record during the EM step posterior counts $c(W)$ for each
composed word pair $W = w_1\circ w_2$.

The effect on the description length of adding a new word can not be
exactly computed.  Its addition will not only affect the counts of other
words, but may also cause other words to be added or deleted.  Fortunately,
simple approximations of the change are adequate for evaluating word
candidates. For example, if Viterbi analyses are being used then the new
word $W$ (if worth adding at all) will completely replace all compositions
of $w_1$ and $w_2$, though each of these words will be used once in the
representation of $W$.  Therefore, if $c(w)$ is the count of a word $w$
before $W$ is added to the lexicon, and $c'(w)$ the count after, then under
the assumption that otherwise parses are stable across the change, $c'(W) =
c(W)$, $c'(w_1) = c(w_1) - c(W) + 1$, $c'(w_2) = c(w_2) - c(W) + 1$ and
otherwise $c'(w) = c(w)$.  Of course, all word probabilities change because
of the change in total word count.  Since the codelength of a word $w$ with
probability $p(w)$ is approximately $-\log p(w)$, the estimated total
change in description length caused by adding a new word $W$ to a lexicon
$L$ is

\begin{eqnarray*}
\Delta& \approx& -c'(W)\log p'(W)  + \mbox{d.l.(changes)}+\\
&& \sum_{w\in L}(-c'(w)\log p'(w) + c(w)\log p(w))
\end{eqnarray*}

\noindent where \mbox{d.l.(changes)} represents the cost of writing down
the perturbations involved in the representation of
$W$.\footnote{See~\cite{deMarcken95d} for more detailed discussion of
  approximations.  The actual schemes used in the tests discussed in this
  paper are slightly more complicated than those presented here.  For
  example, it is not assumed that the representation of $W$ after the
  change will necessarily be $w_1\circ w_2$ and the possibility that either
  or both of $w_1$ and $w_2$ will subsequently be deleted is considered.
  Further, unless Viterbi analyses are being used, $c'(W)$ is not assumed to
  be exactly $c(W)$.} This can be computed quite efficiently.  If
$\Delta<0$ the word $W$ is predicted to reduce the total description length
and is added to the lexicon.  In our implementation, all candidates with
negative $\Delta$ are added simultaneously; subsequent delete steps can fix
mistakes.

Similar heuristic approximations can be used to estimate the benefit of
deleting words.  In that case, a reasonable assumption is that if a word is
deleted its representation replaces it everywhere.  Again this is not
necessarily correct, but serves adequately.

\subsection{Search Properties}

Local optima debilitate many traditional grammar induction techniques
\cite{deMarcken95b,Pereira92,Carroll92}.  The search algorithm described
above generally escapes this problem, in large part because of the
underlying representation.  The reason is that hidden structure is largely
a ``compile-time'' phenomena.  During parsing all that is important about a
word is its surface form and codelength.  The internal representation does
not matter.  Therefore, the internal representation is free to reorganize
at any time; it has been decoupled.  This allows structure to be built
bottom up or for structure to emerge inside already existing parameters.
Furthermore, since parameters (words) encode surface patterns, their use is
constrained and they tend not have competing roles, in contrast, for
instance, to hidden nodes in neural networks.  And since the number of
parameters is not fixed, when words do start to have multiple conflicting
roles, they can be split with common substructure shared.  Finally, since
add and delete cycles can compensate for initial mistakes, inexact
heuristics can be used for adding and deleting words.

\section{Concatenation Results}\label{concat}

The simplest reasonable instantiation of the composition-and-perturbation
framework is with the concatenation operator and probability perturbation.
This instantiation has been tested on problems of text segmentation and
compression.  Given a text document, the search algorithm tries to find the
lexicon that minimizes total description length.  For testing purposes,
delimiters like spaces and punctuation are removed from the input.  Define
{\em true words} to be minimal character sequences bordered by delimiters
in the original input.  Since the search algorithm parses the input as it
compresses it, it can output the optimal segmentation of the input into
words drawn from the lexicon.  These words are themselves decomposed in the
lexicon, and can be considered to form a tree that terminates in
characters.  This tree can have no more than ${\cal O}(n)$ nodes for an
input of length $n$, even though there are ${\cal O}(n^2)$ possible true
words in such an input; thus, the segmentation tree contains considerable
information.  Define {\em recall} to be the percentage of true words that
occur at some level of the segmentation tree.  Define {\em
  crossing-brackets} to be the percentage of true words that violate the
segmentation tree structure.\footnote{The true word \word{moon} in the
  input \word{the moon} is a crossing-bracket violation of \word{them} in
  the (partial) segmentation tree {\word{[[them][o][on]]}}.}

The algorithm was applied to two texts, a lowercase version of the
million-word Brown corpus with spaces and punctuation removed, and 4
million characters of Chinese news articles in a two-byte/character format.
In the case of the Chinese, which contains no inherent separators like
spaces, segmentation performance is measured relative to another computer
segmentation program that had access to a (human-created) lexicon.  The
algorithm was given the raw encoding and had to deduce the internal
two-byte structure.  In the case of the Brown corpus, word recall was
90.5\% and crossing-brackets was 1.7\%.  For the Chinese word recall was
96.9\% and crossing-brackets was 1.3\%.  In the case of both English and
Chinese, most of the recall violations were words that occurred only once in
the corpus.  Thus, the algorithm did an extremely good job of learning
words and properly using them to segment the input.  Furthermore, the
crossing-bracket measure indicates that the algorithm makes very few
clear mistakes.  Of course, the hierarchical lexical representation does
not make a commitment to what levels are ``true words'' and which are not;
about five times more nodes exist in the segmentation tree than true words.
Experiments in section~\ref{meaning} demonstrate that for most applications
this excess structure is not only not a problem, but desirable.
Figure~\ref{fig:brown} displays some of the lexicon learned from the Brown
corpus.

\begin{figure}[th]
\pageline
\begin{small}
\begin{center}
\begin{tabular}{rl}
Rank & Word \\ \hline
    0& \verb|[s]|\\
    1& \verb|[the]|\\
    2& \verb|[and]|\\
    3& \verb|[a]|\\
    4& \verb|[of]|\\
    5& \verb|[in]|\\
    6& \verb|[to]|\\
  500& \verb|[students]|\\
  501& \verb|[material]|\\
  502& \verb|[um]|\\
  503& \verb|[words]|\\
  504& \verb|[period]|\\
  505& \verb|[class]|\\
  506& \verb|[question]|\\
 5000&   \verb|[[ing][them]]|\\
 5001&   \verb|[[mon][k]]|\\
 5002&   \verb|[[re][lax]]|\\
 5003&   \verb|[[rig][id]]|\\
 5004&   \verb|[[connect][ed]]|\\
 5005&   \verb|[[i][k]]|\\
 5006&   \verb|[[hu][t]]|\\
26000&   \verb|[[pleural][blood][supply]]|\\
26001&   \verb|[[anordinary][happy][family]]|\\
26002&   \verb|[[f][eas][ibility][of]]|\\
26003&   \verb|[[lunar][brightness][distribution]]|\\
26004&   \verb|[[primarily][diff][using]]|\\
26005&   \verb|[[sodium][tri][polyphosphate]]|\\
26006&   \verb|[[charcoal][broil][ed]]|\\
\end{tabular}
\end{center}
\end{small}
\caption{\label{fig:brown}Sections of a 26,027 word lexicon learned from
  the Brown corpus, ranked by frequency.  The words in the less-frequent
  half are listed with their first-level decomposition.  Word 5000 causes
  crossing-bracket violations, and words 26002 and 26006 have internal
  structure that causes recall violations.} \pageline
\end{figure}

The algorithm was also run as a compressor on a lower-case version of the
Brown corpus with spaces and punctuation left in.  All bits necessary for
exactly reproducing the input were counted.  Compression performance is
2.12 bits/char, significantly lower than popular algorithms like {\em gzip}
(2.95 bits/char).  This is the best text compression result on this corpus
that we are aware of, and should not be confused with lower
figures~\cite{Brown92} that do not include the cost of parameters.
Furthermore, because the compressed text is stored in terms of linguistic
units like words, it can be searched, indexed, and parsed without
decompression.

\section{Learning Meanings}\label{meaning}

Unsupervised learning algorithms are rarely used in isolation.  The goal of
this work has been to explain how linguistic units like words can be
learned, so that other processes can make use of these units.  In this
section a means of learning the mappings between words and artificial
representations of meanings is described.  The composition-and-perturbation
representation handles this application neatly.

Imagine that text utterances are paired with representations of
meaning,\footnote{This framework is easily extended to handle multiple
  ambiguous meanings (with and without priors) and noise, but these
  extensions are not discussed here.} and that the goal is to find the
minimum-length description of both the text and the meaning.  If there is
mutual information between the meaning and text portions of the input, then
better compression is achieved if the two streams are compressed
simultaneously than independently.  If a text word has an associated
meaning, then writing down that word to account for some portion of text
also accounts for some portion of the meaning of that text.  The remaining
meaning can be written down more succinctly.  Thus, there is an incentive
to associate meaning with sound, although of course the association pays a
price in the description of the lexicon.

Although it is obviously a naive simplification, many of the interesting
properties of the compositional representation surface even when meanings
are treating as sets of arbitrary symbols.  A word is now both a character
sequence and a set of meaning symbols.  The composition operator
concatenates the characters of its operands and takes the union of their
meaning symbols.  Of course, there must be some way to perturb the default
meaning of a word.  One way to do this is to explicitly write out any
symbols that are present in the word's meaning but not in its components,
or {\em vice versa}.  Thus, the word \sword{red}{RED} might be represented
as \word{r} $\circ$ \word{e} $\circ$ \word{d}+{\sc RED}.  Given an existing
word \sword{berry}{BERRY}, the red berry \sword{cranberry}{RED BERRY} can
be represented \word{c} $\circ$ \word{r} $\circ$ \word{a} $\circ$ \word{n}
$\circ$ \sword{berry}{BERRY}+{\sc RED}.

\subsection{Results}

To test the algorithm's ability to infer word meanings, 10,000 utterances
from an unsegmented textual database of mothers' speech to children were
paired with representations of meaning, constructed by assigning a unique
symbol to each root word in the vocabulary.  For example, the sentence {\em
  andwhatishepaintingapictureof} is paired with the unordered meaning $\{$
{\sc AND WHAT BE HE PAINT A PICTURE OF} $\}$.\footnote{The unordered nature
  of the second data stream greatly increases the complexity of the EM
  algorithm, which can no longer be implemented efficiently through dynamic
  programming.  Although too complex to be discussed here, in our
  implementation a factorial approximation is used to succinctly and
  efficiently represent forward and backward probabilities.} In the first
experiment, the algorithm received these pairs with no noise or ambiguity,
using a perturbation operator such that each symbol's cost was 10 bits.
After 8 iterations of training on the text portion of the input and then a
further 8 iterations of training on both the text and the meaning, the text
was parsed again.  The meanings of the resulting word sequences (as defined
by the lexicon) were compared with the true meaning of the input.  Symbol
accuracy was 98.9\%, recall was 93.6\%.  Used to identify the true meaning
from among the meanings of the previous 20 sentences, the program selected
correctly 89.1\% of the time, or ranked the true meaning tied for first
10.8\% of the time.

A second test was performed in which during training the algorithm received
three possible meanings for each utterance, the true one and also the
meanings of the two surrounding utterances.  A uniform prior was used.
Despite the ambiguity, during testing symbol accuracy was again 98.9\%,
recall was 75.3\%.

The final lexicon includes extended phrases, but meanings tend to filter
down to the proper level.  For instance, although the words \word{duck},
\word{ducks}, \word{theducks} and \word{duckdrink} are all in the lexicon
and contain the meaning {\sc DUCK}, the symbol is only written once, in the
description of \word{duck}.  All others words inherit the symbol from this
word.  Similar results hold for similar experiments on the Brown corpus.
For example, \word{scratching her nose} inherits its meaning completely
from its parts, while \word{kicking the bucket} does not.  This is exactly
the result argued for in the motivation section of this paper, and
illustrates why in our framework there is little harm in occasionally
adding unnecessary words like \word{scratching her nose} to the lexicon.

\section{Other Extensions}

We have performed other experiments using this representation and search
algorithm, on tasks in unsupervised learning from speech and grammar
induction.  Figure~\ref{fig:speech} contains a small portion of a lexicon
learned from 55,000 utterances of continuous speech by multiple speakers.
The utterances are taken from dictated Wall Street Journal articles.  The
concatenation operator was used with phonemes as terminals.  A second layer
was added to the framework to map from phonemes to speech; these extensions
are described in more detail in \cite{deMarcken95d}.  The sound models for
the phonemes were estimated independently on a separate corpus of
hand-segmented speech.  Although the phoneme models are extremely poor,
many words are recognizable, and this is the first significant lexicon
learned directly from spoken speech without supervision.

\begin{figure}[h]
\pageline
\def\ipa{\ipatenrm}
\begin{center}
\begin{tabular}{rll}
Rank & $w$ & $\mbox{rep}(w)$\\ \hline 
 5392&[\uniw\uniO\unir\unim\unir]&  [[\uniw\uniO\unir]\unim\unir] \\
 5393&[\uniT\unia\uniu\uniz\unin]&  [\uniT[\unia\uniu\uniz\unin]] \\
 5394&[\unit\uniAX\unih\uniI\unid]&  [[\unit\uniAX\unih]\uniI\unid] \\
 5395&[\uniE\unik\unit\uniI\unid]&  [\uniE\unik[\unit\uniI\unid]] \\
 5396&[\uniAH\unin\unii\uniIX\unin]&  [\uniAH\unin[\unii\uniIX\unin]] \\
 5397&[\unim\uniE\unil\unii\uniIX\unin\unid\unia\unil\unir\uniz]&  [[\unim\uniE\unil\unii\uniIX\unin\unid\unia\unil\unir]\uniz] \\
 8948&[\unia\unii\unid\unii\uniIX\uniz]&  [[\unia\unii]\unid\unii\uniIX\uniz] \\
 8949&[\unis\uniIX\unik\unir\unit\unii]&  [\unis\uniIX\unik[\unir\unit\unii]] \\
 8950&[\unil\uniO\uniG\unit\unia\unii\unim]&  [[\unil\uniO\uniG][\unit\unia\unii\unim]] \\
 8951&[\unis\uniE\unik\unig\uniI\unin]&  [[\unis\uniE\unik][\unig\uniI\unin]] \\
 8952&[\uniw\uniAH\unin\unip\uniAH]&  [[\uniw\uniAH\unin]\unip\uniAH] \\
 8953&[\univ\uniE\unin\unid\uniC\unir]&  [\univ[\uniE\unin][\unid\uniC\unir]] \\
 8954&[\uniAX\unil\uniI\unim\uniI\unin\unie\unii]&  [\uniAX[\unil\uniI\unim\uniI\unin][\unie\unii]] \\
 8955&[\unim\uniE\unil\unii\uniIX\uniG]&  [[\unim\uniE\unil]\unii[\uniIX\uniG]] \\
 8956&[\unib\uniE\unil\unii\uniIX\unin\unid\unia\unil]&  [\unib\uniE[\unil\unii\uniIX\unin\unid\unia\unil]] \\
 9164&[\unig\unio\uniu\unil\unid\unim\uniIX\unin\unis\uniA\unik\unis]&  [[\unig\unio\uniu\unil]\unid[\unim\uniIX\unin]\unis[\uniA\unik\unis]] \\
 9165&[\unik\unim\unip\uniS\uniu\unit\unir]&  [[\unik\unim\unip][\uniS\uniu\unit]\unir] \\
 9166&[\unig\unia\univ\unir\unim\uniIX\unin]&  [\unig\unia[\univ\unir\unim\uniIX\unin]] \\
 9167&[\unio\uniu\unib\unil\uniz\uniAX\unih\uniu\unio\uniu]&  [[\unio\uniu\unib\unil][\uniz\uniAX\unih\uniu\unio\uniu]] \\
 9168&[\unim\uniIX\unin\uniIX\unis\unit\unir\unie\unii\uniS\uniIX\unin]&  [[\unim\uniIX\unin]\uniIX[\unis\unit\unir\unie\unii\uniS\uniIX\unin]] \\
 9169&[\unit\uniJ\uniE\unir\uniIX\unin]&  [[\unit\uniJ\uniE]\unir[\uniIX\unin]] \\
 9170&[\unih\uniAH\unib\unil\unih\uniAX\unih\uniw\unio\uniu]&  [[\unih\uniAH\unib\unil][\unih\uniAX\unih\uniw\unio\uniu]] \\
 9171&[\unis\uniAH\unim\unip\uniD\uniIX\uniG]&  [\unis[\uniAH\unim\unip][\uniD\uniIX\uniG]] \\
 9172&[\unip\unir\unip\unil\unio\uniu\uniz\unil]&  [[\unip\unir][\unip\unil\unio\uniu]\uniz\unil] \\
 9173&[\unib\unio\uniu\unis\unik\unig\unii]&  [[\unib\unio\uniu][\unis\unik\unig]\unii] \\
 9174&[\unik\unig\uniE\unid\uniJ\uniIX\unil]&  [[\unik\unig\uniE][\unid\uniJ\uniIX]\unil] \\
 9175&[\unig\unio\uniu\unil\unid\unim\unia\unii\uniIX\unin\uniz]&  [[\unig\unio\uniu\unil]\unid[\unim\unia\unii\uniIX\unin\uniz]] \\
 9176&[\unik\uniO\unir\unip\unir\unie\unii\unit\uniI\unid]&
 [[\unik\uniO\unir\unip\unir][\unie\unii\unit\uniI\unid]] \\
\end{tabular}
\end{center}
\caption{\label{fig:speech}Some words from a lexicon learned from 55,000
  utterances of continuous, dictated Wall Street Journal articles.
  Although many words are little more than random gibberish, words representing
  \word{million dollars}, \word{Goldman-Sachs}, \word{thousand}, etc.\ are
  learned.  Furthermore, as word 8950 (\word{long time}) demonstrates, they are
  often properly decomposed into components.}
\pageline
\end{figure}
\label{grammar}

If the composition operator makes use of context, then this framework
extends naturally to a variation of stochastic context-free grammars in
which composition corresponds to tree substitution and the inside-outside
algorithm~\cite{Baker79} is used for re-estimation.  In particular, if
each word is associated with a parent class, and these classes are
permissible terminals, then ``words'' act as production rules. For example,
a possible word with class {\em vp} is \verb|[|$_{\mbox{\em vp}}$\mbox{\em
 take off}\verb|<|{\em np}\verb|>]|, which can be represented by
\verb|[|$_{\mbox{\em vp}}$\verb|<|{\em v}\verb|><|{\em p}\verb|><|{\em
 np}\verb|>]|$\circ$\verb|[|$_{\mbox{\em v}}${\em
 take}\verb|]|$\circ$\verb|[|$_{\mbox{\em p}}${\em
 off}\verb|]|$\circ$\verb|[|$_{\mbox{\em np}}\diamond$\verb|]| where
$\diamond$ is a special symbol that indicates a class is not expanded.
Furthermore, \verb|[|$_{\mbox{\em vp}}$\verb|<|{\em v}\verb|>|\verb|<|{\em
 p}\verb|><|{\em np}\verb|>]| may be decomposed into
 \verb|[|$_{\mbox{\em vp}}$\verb|<|{\em v}\verb|><|{\em
 pp}\verb|>]|$\circ$\verb|[|$_{\mbox{\em
 v}}\diamond$\verb|]|$\circ$\verb|[|$_{\mbox{\em pp}}$\verb|<|{\em
 p}\verb|><|{\em np}\verb|>]|.  In this way syntactic structure emerges in
the internal representation of relatively flat production rules.  This
framework offers the significant advantage that non-independent rule
expansions can be accounted for without sacrificing structure.  We are
currently looking at various methods for automatically acquiring classes;
in initial experiments some of the first classes learned from text are the
class of vowels, of consonants, and of verb endings.  

\section{Conclusions}

No previous unsupervised language-learning procedure has produced
structures that match so closely with linguistic intuitions.  We take this
as a vindication of the perturbation-of-compositions representation.  Its
ability to capture the statistical and linguistic idiosyncrasies of large
structures without sacrificing the obvious regularities within them makes
it a valuable tool for a wide variety of induction problems.

This research was supported in part by NSF grant 9217041-ASC and ARPA under
the HPCC and AASERT programs.

\begin{small}

\bibliographystyle{acl}

\begin{thebibliography}{}

\bibitem[\protect\citename{Baker}1979]{Baker79}
J.~K. Baker.
\newblock 1979.
\newblock Trainable grammars for speech recognition.
\newblock In {\em Proceedings of the 97th Meeting of the Acoustical Society of
  America}, pages 547--550.

\bibitem[\protect\citename{Baum \bgroup et al.\egroup }1970]{Baum70}
L.~E. Baum, T. Petrie, G. Soules, and N. Weiss.
\newblock 1970.
\newblock A maximization technique occuring in the statistical analysis of
  probabilistic functions in {M}arkov chains.
\newblock {\em Annals of Mathematical Statistics}, 41:164--171.

\bibitem[\protect\citename{Brown \bgroup et al.\egroup }1992]{Brown92}
P.~L. Brown, S.~A. Della Pietra, V.~J. Della Pietra, J.~C. Lai, and R.~L. Mercer.
\newblock 1992.
\newblock An estimate of an upper bound for the entropy of english.
\newblock {\em Computational Linguistics}, 18(1):31--40.

\bibitem[\protect\citename{Carroll and Charniak}1992]{Carroll92}
G. Carroll and E. Charniak.
\newblock 1992.
\newblock Learning probabilistic dependency grammars from labeled text.
\newblock In {\em Working Notes, Fall Symposium Series, AAAI}, pages 25--31.

\bibitem[\protect\citename{Cartwright and Brent}1994]{Cartwright94}
T.~A. Cartwright and M.~R. Brent.
\newblock 1994.
\newblock Segmenting speech without a lexicon: Evidence for a bootstrapping
  model of lexical acquisition.
\newblock In {\em Proc. of the 16th Annual Meeting of the Cognitive Science
  Society}, Hillsdale, New Jersey.

\bibitem[\protect\citename{Chen}1995]{Chen95}
S.~F. Chen.
\newblock 1995.
\newblock Bayesian grammar induction for language modeling.
\newblock In {\em Proc. 32nd Annual Meeting of the Association for
  Computational Linguistics}, pages 228--235, Cambridge, Massachusetts.

\bibitem[\protect\citename{Chomsky}1955]{Chomsky55}
N.~A. Chomsky.
\newblock 1955.
\newblock {\em The Logical Structure of Linguistic Theory}.
\newblock Plenum Press, New York.

\bibitem[\protect\citename{de Marcken}1995a]{deMarcken95b}
C. de~Marcken.
\newblock 1995a.
\newblock Lexical heads, phrase structure and the induction of grammar.
\newblock In {\em Third Workshop on Very Large Corpora}, Cambridge,
  Massachusetts.

\bibitem[\protect\citename{de Marcken}1995b]{deMarcken95d}
C. de~Marcken.
\newblock 1995b.
\newblock The unsupervised acquisition of a lexicon from continuous speech.
\newblock Memo A.I.~Memo 1558, MIT Artificial Intelligence Lab., Cambridge,
  Massachusetts.

\bibitem[\protect\citename{Deligne and Bimbot}1995]{Deligne95}
S. Deligne and F. Bimbot.
\newblock 1995.
\newblock Language modeling by variable length sequences: Theoretical
  formulation and evaluation of multigrams.
\newblock In {\em Proceedings of the International Conference on Speech and
  Signal Processing}, volume~1, pages 169--172.

\bibitem[\protect\citename{Dempster \bgroup et al.\egroup }1977]{Dempster77}
A.~P. Dempster, N.~M. Liard, and D.~B. Rubin.
\newblock 1977.
\newblock Maximum liklihood from incomplete data via the {EM} algorithm.
\newblock {\em Journal of the Royal Statistical Society}, B(39):1--38.

\bibitem[\protect\citename{Ellison}1992]{Ellison92}
T.~M. Ellison.
\newblock 1992.
\newblock {\em The Machine Learning of Phonological Structure}.
\newblock {Ph.D.} thesis, University of Western Australia.

\bibitem[\protect\citename{Francis and Kucera}1982]{Francis82}
W.~N. Francis and H.~Kucera.
\newblock 1982.
\newblock {\em Frequency analysis of English usage: lexicon and grammar}.
\newblock Houghton-Mifflin, Boston.

\bibitem[\protect\citename{Harris}1968]{Harris68}
Z. Harris.
\newblock 1968.
\newblock {\em Mathematical Structure of Language}.
\newblock Wiley, New York.

\bibitem[\protect\citename{Olivier}1968]{Olivier68}
D.~C. Olivier.
\newblock 1968.
\newblock {\em Stochastic Grammars and Language Acquisition Mechanisms}.
\newblock {Ph.D.} thesis, Harvard University, Cambridge, Massachusetts.

\bibitem[\protect\citename{Pereira and Schabes}1992]{Pereira92}
F. Pereira and Y. Schabes.
\newblock 1992.
\newblock Inside-outside reestimation from partially bracketed corpora.
\newblock In {\em Proc. 29th Annual Meeting of the Association for
  Computational Linguistics}, pages 128--135, Berkeley, California.

\bibitem[\protect\citename{Rissanen}1978]{Rissanen78}
J. Rissanen.
\newblock 1978.
\newblock Modeling by shortest data description.
\newblock {\em Automatica}, 14:465--471.

\bibitem[\protect\citename{Rissanen}1989]{Rissanen89}
J. Rissanen.
\newblock 1989.
\newblock {\em Stochastic Complexity in Statistical Inquiry}.
\newblock World Scientific, Singapore.

\bibitem[\protect\citename{Solomonoff}1960]{Solomonoff60}
R.~J. Solomonoff.
\newblock 1960.
\newblock The mechanization of linguistic learning.
\newblock In {\em Proceedings of the 2nd International Conference on
  Cybernetics}, pages 180--193.

\bibitem[\protect\citename{Stolcke}1994]{Stolcke94}
A. Stolcke.
\newblock 1994.
\newblock {\em Bayesian Learning of Probabilistic Language Models}.
\newblock {Ph.D.} thesis, University of California at Berkeley, Berkeley, CA.

\bibitem[\protect\citename{Wolff}1982]{Wolff82}
J.~G. Wolff.
\newblock 1982.
\newblock Language acquisition, data compression and generalization.
\newblock {\em Language and Communication}, 2(1):57--89.

\bibitem[\protect\citename{Ziv and Lempel}1978]{Ziv78}
J.~Ziv and A.~Lempel.
\newblock 1978.
\newblock Compression of individual sequences by variable rate coding.
\newblock {\em IEEE Transactions on Information Theory}, 24:530--536.

\end{thebibliography}

\end{small}
\end{document}